\documentclass[lettersize,journal]{IEEEtran}
\usepackage{algorithmic}
\usepackage{array}
\usepackage[caption=false,font=normalsize,labelfont=sf,textfont=sf]{subfig}
\usepackage{textcomp}
\usepackage{stfloats}
\usepackage{url}
\usepackage{verbatim}
\usepackage{graphicx}
\def\BibTeX{{\rm B\kern-.05em{\sc i\kern-.025em b}\kern-.08em
    T\kern-.1667em\lower.7ex\hbox{E}\kern-.125emX}}
\usepackage{balance}
\usepackage{cite}
\usepackage{amsmath,amssymb,amsfonts}
\usepackage{xcolor}

\usepackage{booktabs}
\newtheorem{definition}{Definition}
\usepackage{algorithm}
\usepackage{newfloat}
\usepackage{listings}
\usepackage{xspace} 
\usepackage{multirow}
\usepackage{pifont}
\usepackage{tabularx}

\newcommand{\etal}{{\it et al.}\xspace}

\begin{document}

\title{From Conflicts to Collisions: A Two-Stage Collision Scenario-Testing Approach for Autonomous Driving Systems}


\author{Siyuan~Chen,
        Fuyuan~Zhang,
        Hua~Qi,
        Lei~Ma,
        Tomoyuki~Tsuchiya,
        Michio~Hayashi,
        Manabu~Okada
\IEEEcompsocitemizethanks{\IEEEcompsocthanksitem S. Chen is with The University of Tokyo, Tokyo, Japan. \protect\\
E-mail: alsachai@g.ecc.u-tokyo.ac.jp
\IEEEcompsocthanksitem F. Zhang is with Zhejiang University, Hangzhou, China.
\IEEEcompsocthanksitem H. Qi is with Kyushu University, Fukuoka, Japan, and also with The University of Tokyo, Tokyo, Japan.
\IEEEcompsocthanksitem L. Ma is with The University of Tokyo, Tokyo, Japan, and also with the University of Alberta, Edmonton, Canada.
\IEEEcompsocthanksitem T. Tsuchiya and M. Okada are with TIER IV, Tokyo, Japan.
\IEEEcompsocthanksitem M. Hayashi is with TIER IV North America, Palo Alto, CA, USA.}%
\thanks{Corresponding author: Lei Ma (e-mail: ma.lei@acm.org).}}

\maketitle

\begin{abstract}
Autonomous driving systems (ADS) are safety-critical and require rigorous testing before public deployment. Simulation-based scenario testing provides a safe and cost-effective alternative to extensive on-road trials, enabling efficient evaluation of ADS under diverse and high-risk conditions. However, existing approaches mainly evaluates the scenarios based on their proximity to collisions and focus on scenarios already close to collision, leaving many other hazardous situations unexplored. To bridge this, we introduce a collision-related concept of conflict as an intermediate search target and propose a two-stage scenario testing framework that first searches for conflicts and then mutates these conflict scenarios to induce actual collisions. Evaluated on Baidu Apollo, our approach reveals up to 12 distinct collision types in a single run, doubling the diversity discovered by state‑of‑the‑art baselines while requiring fewer simulations thanks to conflict‑targeted mutations. These results show that using conflicts as intermediate objectives broadens the search horizon and significantly improves the efficiency and effectiveness of ADS safety evaluation.
\end{abstract}


\section{Introduction}

Autonomous Driving Systems (ADSs) are poised to revolutionize transportation by reducing human error and mitigating traffic congestion, yet their safety-critical nature demands rigorous testing before public deployment~\cite{DVN/MCNENT_2020}. While real-world validation involves extensive on-road driving and high costs, simulation-based testing~\cite{Arrieta2017test,kim2019test,feng2022test,yan2022naturalistic}, which can expose ADSs to diverse and high-risk conditions without endangering human lives or physical vehicles, offers a more efficient alternative for evaluating ADS performance.

One key branch of simulation-based testing is scenario-based testing~\cite{neurohr2020fundamental}, which constructs scenarios by varying parameters (e.g., traffic participants, road infrastructure) to systematically reveal potential ADS failures. Among the simulation-based testings~\cite{dreossi2019verifai,fremont2020formal,sun2022lawbreaker,han2020metamorphic}, search-based scenario testing methods~\cite{huai2023doppelganger,kim2022drivefuzz,li2020av,lu2021search,tian2022mosat,tang2024legend} have proven effective at uncovering a wide range of failure-inducing conditions within the large input space. For example, AVFuzzer~\cite{li2020av} applies genetic algorithms with atom-level mutations to control non-player vehicle maneuvers and induce collisions caused by ADS faults. MOSAT~\cite{tian2022mosat} builds on this idea by introducing motif-based mutations, which group multiple atom-level mutations alongside multi-objective functions to enhance the genetic algorithm. Legend~\cite{tang2024legend}, on the other hand, proposes a top-down approach that leverages Large Language Models (LLMs) to translate high-level functional scenarios, often derived from real-world accident reports, into formal logical scenarios for search-based testing.

\begin{figure}[h]
    \centering
    \includegraphics[width=\linewidth]{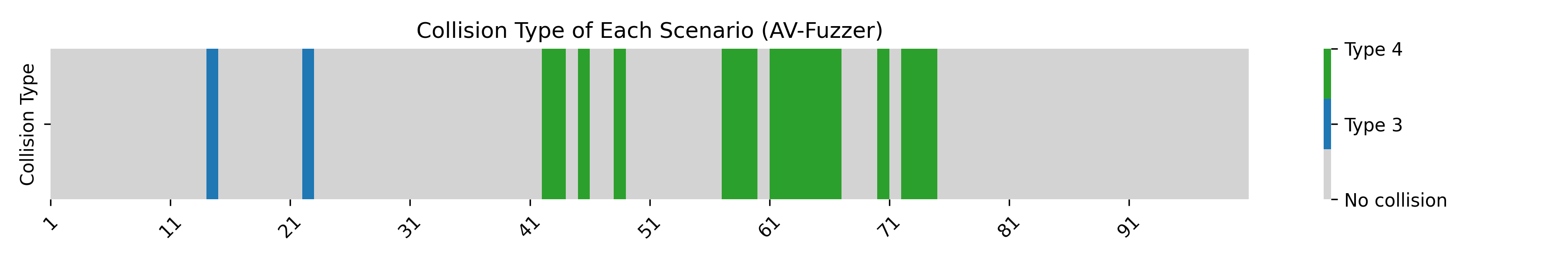} 
    \caption{A visualization of collision types found by AV-Fuzzer over 100 consecutive search steps. Each vertical bar on the chart represents one search step. The gray background indicates a scenario where no collision occurred, while the colored bars highlight a successful collision scenario. Different colors correspond to the distinct collision types detailed in Appendix A.2. The chart shows that within this interval, the collisions found are of limited variety.}
    \label{fig:intro}
\end{figure}

Despite their strengths, previous search-based methods~\cite{li2020av,lu2021search,tian2022mosat,kim2022drivefuzz,tang2024legend} typically evaluate scenarios based on their proximity to collisions, for instance, by measuring the distance between two vehicles. While intuitive, this strategy has a significant limitation: it biases the search toward a narrow subset of near-collision scenarios, overlooking a wide range of other hazardous situations that could evolve into collisions. For example, Figure~\ref{fig:intro} provides an analysis of AV-Fuzzer~\cite{li2020av}, showing the collision types observed over 100 consecutive search steps. It is clear that while AV-Fuzzer successfully finds some collisions, they predominantly belong to the same type. This premature convergence occurs because the genetic algorithm repeatedly selects a small and homogeneous subset of scenarios that yield high distance-based fitness scores to serve as parents for the next generation, while ignoring other potentially hazardous scenarios that are not yet in extremely close proximity to the ego vehicle. As a result, this strategy drives AV-Fuzzer toward the same near-collision pattern, increasing the total collision count without improving the diversity of failures found.

Rather than directly searching for the perfect timing or location that defines a rare collision, we relax this strict condition for better exploration. Specifically, we target the \textbf{conflict}, which is a much more common situation that represents the precursor of a potential collision, to explore various near-collision scenarios. To balance this broad exploration with effective exploitation for collisions, our key insight is a new two-stage framework that explicitly separates the search for diverse scenarios from the final search for collisions by introducing \textbf{conflicts} as an intermediate target. We first employ a broad conflict search to identify a rich set of varied, hazardous interactions. Then, using various near-collision conflicts as high-potential starting points, we initiate a focused collision search. This second stage can effectively leverage proximity-based fitness functions to efficiently convert a specific conflict into a safety-critical collision. This approach overcomes the exploration limits of prior work while still capitalizing on their effective exploitation strategies.

To verify our idea, our framework builds upon the foundational genetic algorithm-based architecture of AV-Fuzzer~\cite{li2020av}, restructuring the search process into two distinct stages: \textit{conflict search} and \textit{collision search}. The conflict search stage still applies a genetic algorithm, guided by a conflict-based fitness function, to efficiently explore and generate a diverse population of conflict scenarios. Subsequently, the collision search stage employs mutation-based fuzzing to transform these safety-critical conflicts into actual collisions. Our approach also employs atom-level conflict-targeted mutations to strategically perturb promising regions of a scenario and our ablation study in Section~\ref{exp:rq3} demonstrates the advantage of these mutations over random strategies. We validate our framework through experiments on Baidu Apollo~\cite{ap}, an industry-grade ADS, comparing it against three state-of-the-art baselines: AV-Fuzzer~\cite{li2020av}, MOSAT~\cite{tian2022mosat}, and Legend~\cite{tang2024legend}. In a straight-road-limited experiment, our method reveals up to 12 distinct collision types, significantly outperforming the 6 and 10 types found by AV-Fuzzer and MOSAT, respectively. This strong performance extends to unconstrained environments, where our approach discovers 14 collision types to Legend's 11. Furthermore, our framework can detect multiple collision types originating from different conflicts within a single scenario, demonstrating its capacity to thoroughly explore a wide range of high-risk conditions.

In conclusion, our paper makes following contributions:
\begin{itemize}
    \item We introduce a novel two-stage scenario testing framework consisting of \textit{conflict search} (targeting conflicts) and \textit{collision search} (targeting collisions). By shifting the focus from near-collision scenarios to conflict scenarios with a new conflict-based fitness function during conflict search, this framework broadens the exploration space for potential collisions, thereby improving effectiveness.
    \item We design atom-level conflict-targeted mutation operations to enhance scenario perturbations. An ablation study demonstrates the superiority of these operators compared to random mutations.
    \item We implement our method on Baidu Apollo, showing that incorporating conflicts significantly improves effectiveness, leading to the discovery of a greater variety of collision scenarios compared to the state-of-the-art baselines.
\end{itemize}

\section{Conflict-based Scenario Testing}

\subsection{Space-sharing Conflicts}
\label{back:conflict}

\begin{figure*}[h]
    \centering
    \includegraphics[width=\linewidth]{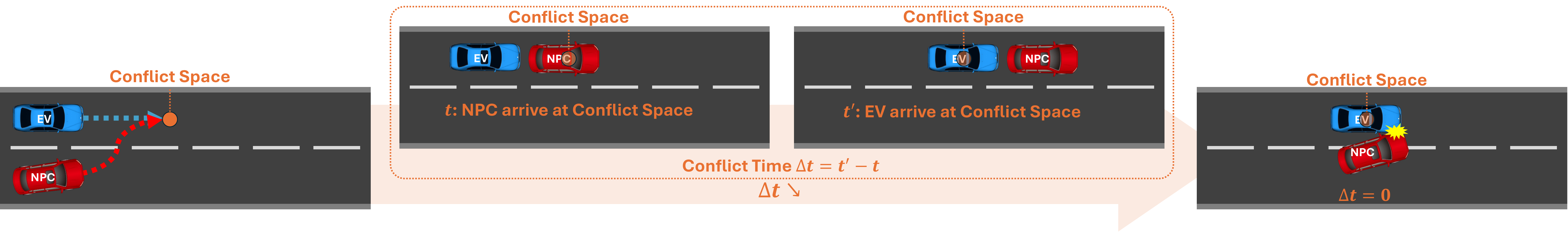}
    \caption{An example of merging path conflict.}
    \label{fig:conflict_example}
\end{figure*}

In the context of autonomous driving and human road interactions, space-sharing conflicts play a central role in understanding how road users behave when they intend to occupy the same region of space. A conflict arises when two or more road users attempt to occupy the same location simultaneously or within a short interval, potentially resulting in dangerous collisions. Therefore, identifying conflicts is crucial for preventing collisions and enhancing road safety. Formally, we define a conflict as follows:
\begin{definition}
\label{def:conflict}
A \textbf{conflict} only occurs when two vehicles approach the same location within a time difference $\Delta t$ that is less than or equal to a predefined conflict time limit $t_c$ (i.e., $0 \leq \Delta t \leq t_c$). As shown in Figure~\ref{fig:conflict_example}, this location is referred to as the \textbf{conflict space}, and the time difference $\Delta t$ is called the \textbf{conflict time}. Note that $t_c$ should be a small value.
\end{definition}

\begin{figure}[h]
    \centering
    \includegraphics[width=\linewidth]{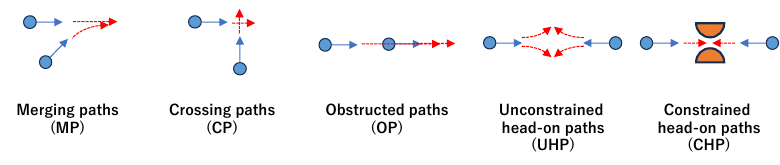}
    \caption{5 types of conflicts.}
    \label{fig:5_conflicts}
\end{figure}

Markkula et al.~\cite{markkula2020defining} proposed five prototypical types of conflicts - 1. Merging Path, 2. Crossing Path, 3. Obstructed Path, 4. Unconstrained Head-On Path and 5. Constrained Head-On Path. We show these five types of conflict in Figure~\ref{fig:5_conflicts} and briefly introduce them in the following.

\textbf{Merging Path (MP).} Merging Path arises at junctions or lane merges when two road users are on paths that will converge into a single one.

\textbf{Crossing Path (CP).} Crossing Path occurs when the paths of two road users intersect, typically at an intersection like a crossroad.

\textbf{Obstructed Path (OP).} Obstructed Path arises when a road user faces an obstacle or another road user blocking their intended path.

\textbf{Unconstrained Head-On Path (UHP).} Unconstrained Head-On Path arises when two road users are approaching each other head-on without any physical constraints.

\textbf{Constrained Head-On Path (CHP).} Constrained Head-On Path occurs when two road users approach each other head-on on a narrow or constrained road, such as a single-lane two-way road.

Figure~\ref{fig:conflict_example} illustrates a typical conflict scenario involving two vehicles: one continuing straight and the other merging, attempting to occupy the same conflict space. When their arrival times at this space differ by only a small interval (conflict time), a merging path conflict occurs. If this interval is zero, a collision occurs, presenting a safety-critical scenario.

\subsection{Using Conflicts for Scenario Testing}
\label{back:testing}

While directly targeting near-collision scenarios is the primary strategy of previous search-based methods, their relative rarity within the expansive scenario space makes them inefficient targets for broad exploration. As discussed in Section~\ref{back:conflict}, conflicts signal collision-related risks and serve as more frequent precursors to crashes. By using conflicts as an initial search target, our framework can first explore a wide range of diverse, hazardous situations before focusing on generating specific collision events.

To leverage this insight, we propose a two-stage framework targeting both conflicts and collisions. As illustrated in Figure~\ref{fig:overview}, our testing framework consists of two main components: \ding{182}\textbf{Conflict Search}, which aims at exploring and diversifying conflicts, and \ding{183}\textbf{Collision Search}, which refines these conflict scenarios into collision scenarios.

The primary goal of the conflict search stage is to maximize the diversity of conflicts within a single scenario. We achieve this by increasing the total number of conflicts, as each one represents a unique interaction based on vehicle positions, speeds, and actions. Therefore, a greater number of conflicts within a single scenario directly leads to broader exploration diversity for the subsequent collision search stage. According to Definition~\ref{def:conflict}, conflicts require meeting both spatial (same location) and temporal ($\Delta t \leq t_c$) criteria. In practice, many situations fulfill only the spatial criterion, for example, two vehicles arriving at the same location separated by a significant time gap. While these situations are not considered actual conflicts, they occur frequently and can potentially be transformed into conflicts. For clarity, we refer to these spatially qualifying cases (with $\Delta t$ larger than $t_c$ but within a larger threshold $t_s$) as \textbf{spatial conflicts}. In Section~\ref{ga}, we detail how we leverage spatial conflicts by employing a genetic algorithm to generate new conflicts involving the vehicle under test, thereby increasing the amount of conflicts and enabling a comprehensive exploration of diverse conflict scenarios.

After several iterations of conflict search, we select the scenario containing the highest number of conflicts for the exploitation-focused collision search stage. Similar to previous search-based methods~\cite{li2020av,tian2022mosat,kim2022drivefuzz}, collision search evaluates scenarios based on proximity to collisions, leveraging the proven effectiveness of this searching strategy. However, instead of applying random modifications to the entire scenario, our collision search directly targets conflicts for modifications. This allows an efficient exploitation of the promising scenarios discovered during the exploration phase.  We implement mutation-based fuzzing with tailored mutation strategies corresponding to specific conflict types (e.g., OP conflicts) present in scenarios. Further details on these mutation strategies are provided in Section~\ref{mf}.

\section{Technical Details}

\subsection{Scenario Representation}
\label{method:scenario}

In this paper, we refer to the vehicle connected to the ADS under test as the ego vehicle (EV), while all other non-ego vehicles are termed non-player characters (NPCs). Our testing framework evaluates the EV within pre-generated scenarios, collecting simulation data on both the EV and NPCs to guide the search process.

\begin{figure}[h]
    \centering
    \includegraphics[width=\linewidth]{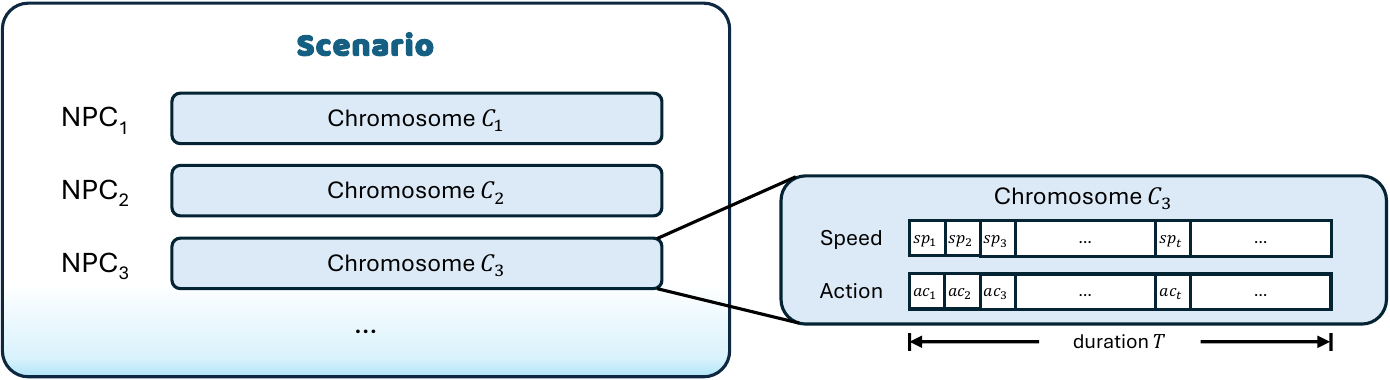}
    \caption{Representation of a scenario.}
    \label{fig:chromosome}
\end{figure}

To facilitate search-based testing, we represent each scenario as a set of chromosomes $S = {C_1, C_2, \dotsc, C_n}$, as illustrated in Figure~\ref{fig:chromosome}, where each chromosome $C_i$ corresponds to the $i^{th}$ NPC in the scenario. Each chromosome spans a scenario duration of $T$ seconds. Specifically, a chromosome $C_i$ for NPC $i$ comprises two gene sequences: a speed series $SP_i$ and an action series $AC_i$. The speed series $SP_i$ defines NPC $i$'s speed at each second within a predefined speed range. The action series $AC_i$ specifies the maneuvers for NPC $i$ at each second, encompassing three fundamental actions: moving straight, changing to the left lane, or changing to the right lane. These two gene series together enable various NPC maneuvers essential for comprehensive scenario testing.

\textbf{Testing Model.} Our testing model assumes no direct control over the target ADS but can control NPC behaviors to effectively challenge the EV.

\subsection{Overview}
\label{method:overview}

\begin{figure*}[h]
    \centering
    \includegraphics[width=\linewidth]{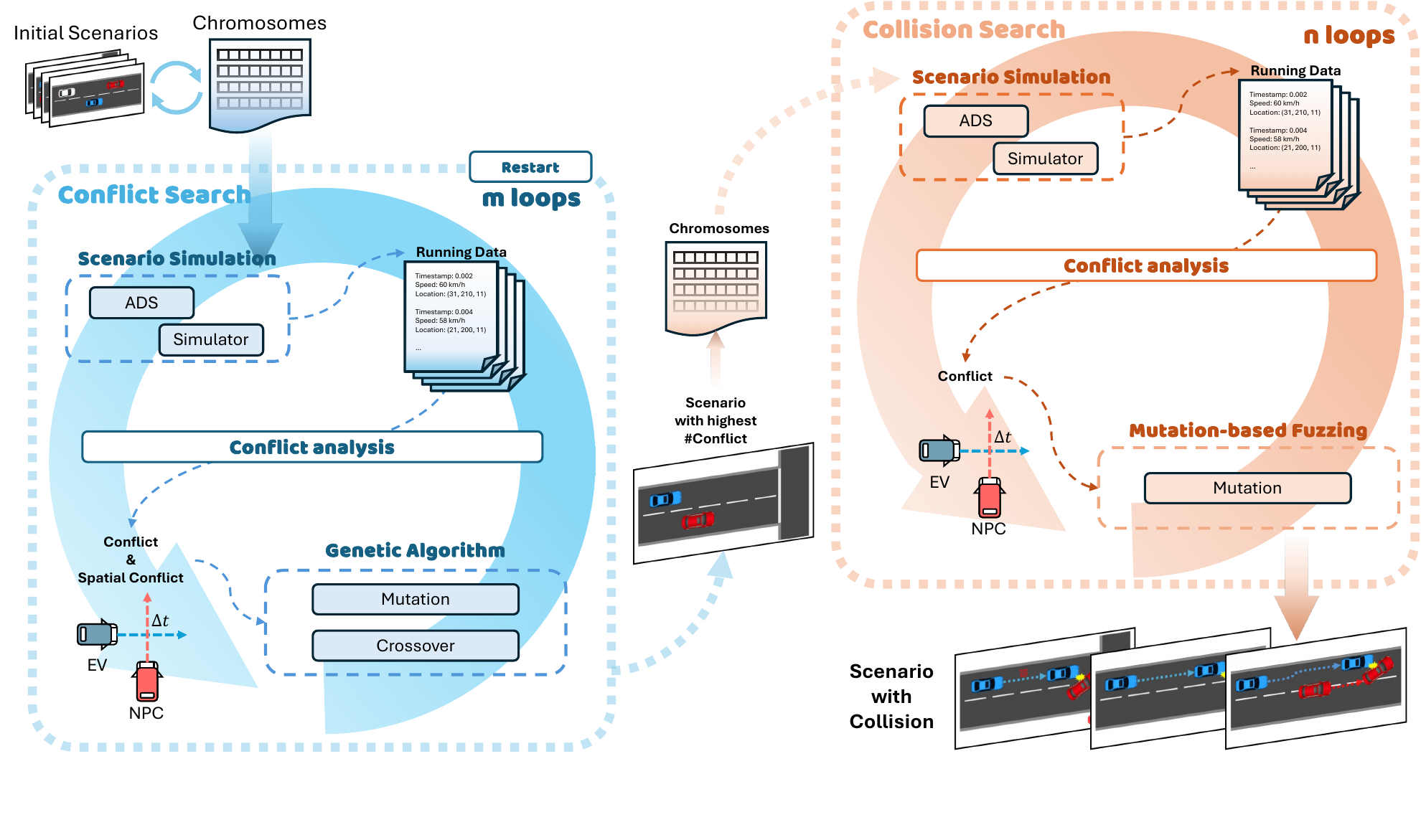}
    \caption{Overview of our approach.}
    \label{fig:overview}
\end{figure*}

Our testing framework adopts the simple genetic algorithm-based architecture from AV-Fuzzer~\cite{li2020av} and Figure~\ref{fig:overview} provides an overview of our approach. The searching process begins by generating an initial population of $k$ randomly initialized scenarios. These scenarios are then used in conflict search, which aims to identify and diversify conflict scenarios. After $m$ iterations of conflict search, the scenario exhibiting the highest number of conflicts is chosen as the starting point for collision search, which iteratively applies mutation-based fuzzing on different conflicts to uncover potential collisions. To further ensure sufficient scenario diversity throughout the search process, we keep the random restart module from AV-Fuzzer. This module periodically evaluates the diversity of previously generated scenarios and generates new initial scenarios for conflict search when existing scenarios become overly similar. Whenever conflict search or collision search detects an EV-caused collision, the corresponding scenario and simulation data are recorded for subsequent replay and detailed analysis.

Both conflict search and collision search contain two primary components: \textit{scenario simulation} and \textit{conflict analysis}. The scenario simulation component takes a chromosome (representing a scenario) as input and executes it within the simulator according to the gene sequences defined in the chromosome. NPCs are initialized at their specified positions and adhere to the speeds and actions encoded in their chromosomes throughout the scenario. The simulation continues until either a collision involving the EV occurs or the scenario duration expires. Throughout the simulation, we continuously record positional and speed data for all NPCs and the EV.

Following each simulation, the conflict analysis component processes the collected simulation data to identify conflicts and spatial conflicts within the executed scenarios. Notably, a single NPC can participate in multiple spatial conflicts and conflicts with the EV, differentiated by distinct locations and times. Conflict search then utilizes these identified conflicts and spatial conflicts to guide the genetic algorithm for exploring new conflict scenarios, while collision search employs mutation-based fuzzing strategies leveraging the identified conflicts to search for potential collisions. In the following sections, we will introduce the details of the genetic algorithm and the mutation-based fuzzing.

\subsection{Genetic Algorithm for Conflict Search}
\label{ga}

Conflict search uses the genetic algorithm to generate scenarios with various conflicts. Specifically, the genetic algorithm takes the initial scenarios as the \textit{population} and iteratively explores new scenarios as the next generation, based on the candidates in the parent population. During the exploration, some variation operations are defined to modify the chromosomes of the scenarios in the population, and a fitness function then evaluates these modified scenarios, retaining the optimized ones for the next generation.

In conflict search, we use two types of variation operations: crossover and mutation. The crossover operation increases population diversity by randomly selecting two scenarios and swapping their chromosomes (NPC settings). Meanwhile, the mutation operations aim to shorten the conflict time of spatial conflicts, thereby creating new conflicts that did not exist in the original scenarios and increasing the total number of conflicts. Since spatial conflicts often have large conflict times, continually applying only small changes would be inefficient. Therefore, we design mutation operations that significantly modify chromosomes to more rapidly convert spatial conflicts into conflicts. These mutation operations are applied to every NPC involved in a spatial conflict with the EV.

\begin{figure}[h]
    \centering
    \includegraphics[width=\linewidth]{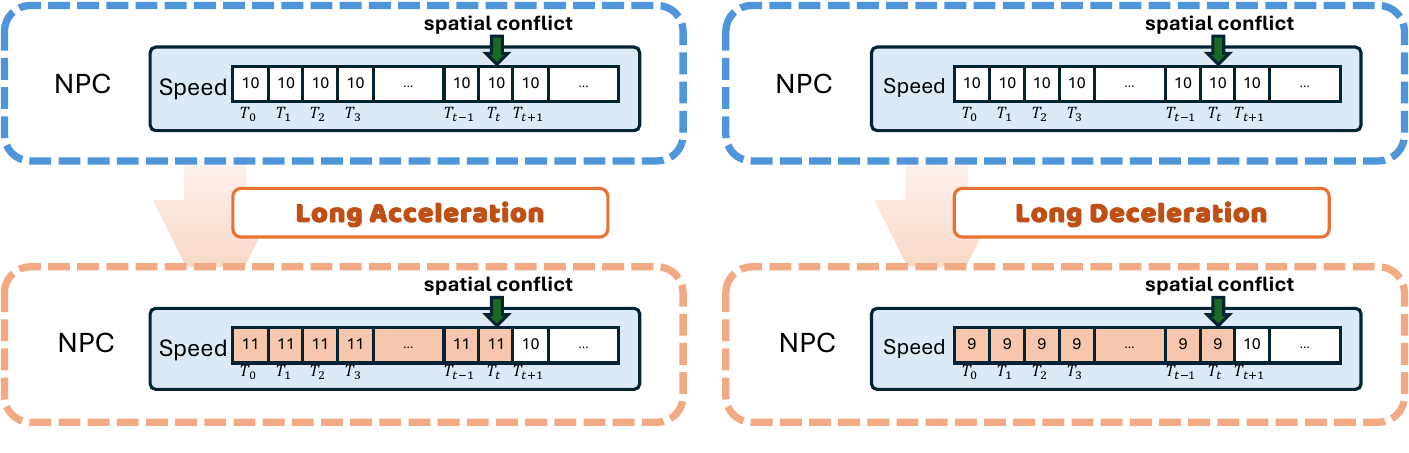}
    \caption{Examples of Long Acceleration and Long Deceleration mutations to perturb the chromosome of an NPC. In both examples, the NPC reaches the spatial conflict space at time $t$.}
    \label{fig:mut_lacc_ldec}
\end{figure}

Assume an NPC $a$ is involved in a group of spatial conflicts $SC_{a} =$ $sc_{a1}$, $sc_{a2}$, $\dotsc$, $sc_{an}$. A spatial conflict $sc_{ab}$ is randomly selected from $SC_{a}$ and NPC $a$ arrives the conflict space of $sc_{ab}$ at time $t$. For NPC $a$, its chromosome will undergo one of two atom-level mutation operations: 
\begin{enumerate}
    \item \textbf{Long Acceleration:} If the EV arrives at the conflict space of $sc_{ab}$ first, increase the speeds of $a$ until it reaches the conflict space, so that $a$ can arrive significantly earlier. The speeds of $a$ will be increased by 1 m/s from the start of the scenario to time $t$, but do not exceed the speed limit. Figure~\ref{fig:mut_lacc_ldec} shows an example of Long Acceleration, where the NPC’s speed increases from 10 m/s to 11 m/s until time $t$.

    \item \textbf{Long Deceleration:} If $a$ arrives at the conflict space of $sc_{ab}$ first, decrease its speeds until it reaches the conflict space so that $a$ effectively waits for the EV to arrive. The speeds of $a$ will be decreased by 1 m/s from the start of the scenario to time $t$, but do not fall below 0 m/s. Figure~\ref{fig:mut_lacc_ldec} shows an example of Long Deceleration, where the NPC’s speed decreases from 10 m/s to 9 m/s until time $t$.
\end{enumerate}

\begin{figure}[h]
    \centering
    \includegraphics[width=\linewidth]{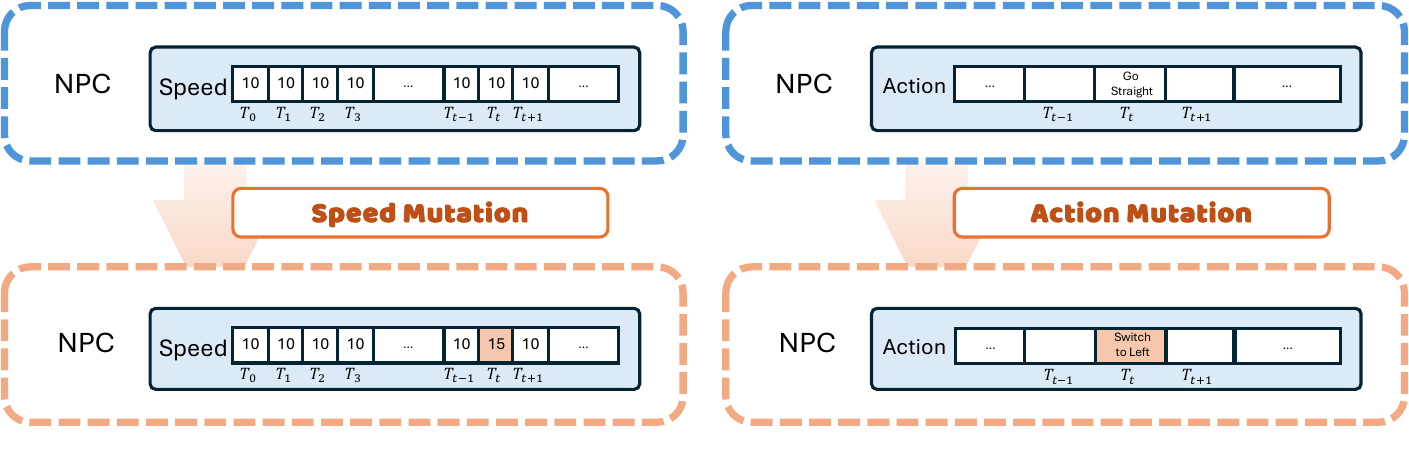}
    \caption{Examples of Speed Mutation and Action Mutation for NPCs not involved in any conflicts or spatial conflicts. In both examples, the time $t$ is a randomly chosen time point within the scenario.}
    \label{fig:mut_spd_act}
\end{figure}

On the other hand, for NPCs not involved in any conflicts or spatial conflicts, random mutations are introduced to create new possible spatial conflicts. These mutations include: 
\begin{enumerate}
    \item \textbf{Speed Mutation:} The speed at a random time is altered within the allowable speed range. Figure~\ref{fig:mut_spd_act} shows an example of Speed Mutation, where the NPC’s speed changes from 10 m/s to 15 m/s at a random time $t$.

    \item \textbf{Action Mutation:} A random action at a random time is changed to a different maneuver, including going straight, switching to the left lane, or switching to the right lane. Figure~\ref{fig:mut_spd_act} shows an example of Action Mutation, where the NPC’s action changes from going straight to switching to the left lane at a random time $t$.
\end{enumerate}

After these variation operations are applied, a fitness function determines which scenarios become parents for the next generation. Since conflict search will output the scenario containing the highest number of conflicts for collision search, we simply design the fitness function as the total number of conflicts in a scenario:
\begin{equation}
\label{eq:conflict}
F_{conflict} = N, \text{$N$ is the number of the conflicts.}
\end{equation}
Scenarios with more conflicts earn higher fitness scores, making them more likely to be selected as parents. By iteratively creating new scenarios through mutation and crossover and then selecting the highest-scoring ones for the next generation, conflict search progressively evolves scenarios toward a higher number of conflicts.

In Algorithm~\ref{algo:conflict}, we show how conflict search works to generate one generation. Conflict search takes the parent population $P$ from last generation and the spatial conflicts set $SC$ of the scenarios from $P$ as inputs, and outputs the next generation $PN$ and spatial conflicts set $SCN$ of $PN$. For each scenario in $P$, conflicts search will randomly generate $r_m$ and $r_c$ respectively and compare with $threshold_m$ and $threshold_c$ to decide whether the scenario will be mutated and crossovered to generate a new scenario (line 3-8). If a new scenario is generated, it will be first simulated with scenario simulation (line 9-16), then analyzed by conflict analysis to find the spatial conflicts and conflicts (line 18-19). The fitness score will be calculated with the Equation~\ref{eq:conflict} (line 20-21) and used for roulette wheel selection of the next generation (line 23). For every $m$ loops, conflict search will output the scenario with the highest fitness score among the past $m$ loops as the input for collision search, and then execute the random restart process.

\begin{algorithm}[tb]
    \caption{Conflict Search}
    \label{algo:conflict}
    \begin{flushleft}
    \textbf{Input}: mutation probability $threshold_m$, crossover probability $threshold_c$,, parent population $P$, spatial conflicts set of parent population $SC$\\
    \textbf{Output}: next population $PN$, spatial conflicts set of next population $SCN$
    \end{flushleft}
    \begin{algorithmic}[1] 
        \STATE $PN \gets \emptyset, S \gets \emptyset, SCN \gets \emptyset.$
        \FOR{i in range(0, $|P|$)}
        \IF{$r_m\sim U(0,1) > threshold_m$}
        \STATE $P^{'}_i$ = $Mutation(P_i, PC_i).$
        \ENDIF
        \IF{$r_c\sim U(0,1) > threshold_c$}
        \STATE $P^{'}_i = Crossover(P^{'}_i, P_j)$, $j \neq i$ and $j \leq |P|$.
        \ENDIF
        \IF{If $P^{'}_i$ existed}
        \STATE Simulate $P^{'}_i$.
        \IF{$P^{'}_i$ has collision}
        \STATE Record Collision.
        \ENDIF
        \ELSE
        \STATE $P^{'}_i = P_i$
        \ENDIF
        \STATE $PN \gets P^{'}_i \cup PN$
        \STATE Find the spatial conflicts $SC^{'}_i$ and conflicts $CFT^{'}_i$ in $P^{'}_i$.
        \STATE $SCN \gets SC^{'}_i \cup SCN$
        \STATE $Score^{'}_i$ = $Fitness(CFT^{'}_i)$
        \STATE $S \gets Score^{'}_i \cup S$ 
        \ENDFOR
        \STATE $PN, SCN = Roulette(PN, SCN, S)$
        \STATE \textbf{return} $PN, SCN$
    \end{algorithmic}
\end{algorithm}

\subsection{Mutation-based Fuzzing for Collision Search}
\label{mf}
Collision search uses mutation-based fuzzing techniques to uncover safety-critical scenarios that result in collisions. Similar to the genetic algorithm in conflict search, collision search proceeds iteratively but focuses exclusively on a single critical scenario to more effectively exploit collisions in that scenario. Starting from the most critical scenario identified in conflict search, collision search initializes its population with that scenario and applies mutation operations repeatedly to induce collisions among various conflicts. During each iteration, if a mutated scenario contains no collisions, a fitness function determines how close that scenario is to producing a collision. Then, the highest-scoring scenario from those newly generated is then chosen as the fuzzing target for the next generation. By iteratively generating and selecting scenarios in this manner, collision search systematically refines conflict scenarios into safety-critical collision scenarios.

As outlined in Section~\ref{back:testing}, we adopt specialized conflict-targeted mutation operations to provide a more targeted and efficient exploration of collisions than purely random modifications. Let $CFT$ denote the conflicts identified in the most critical scenario from conflict search. The mutation target is selected based on one of two criteria: either the conflict with the shortest conflict time (50\% chance) or a randomly selected conflict (50\% chance). For a selected conflict $cft$ occurring at time $t$, the chromosome of the NPC involved in $cft$ is mutated according to the conflict type. 

\begin{figure}[h]
    \centering
    \includegraphics[width=\linewidth]{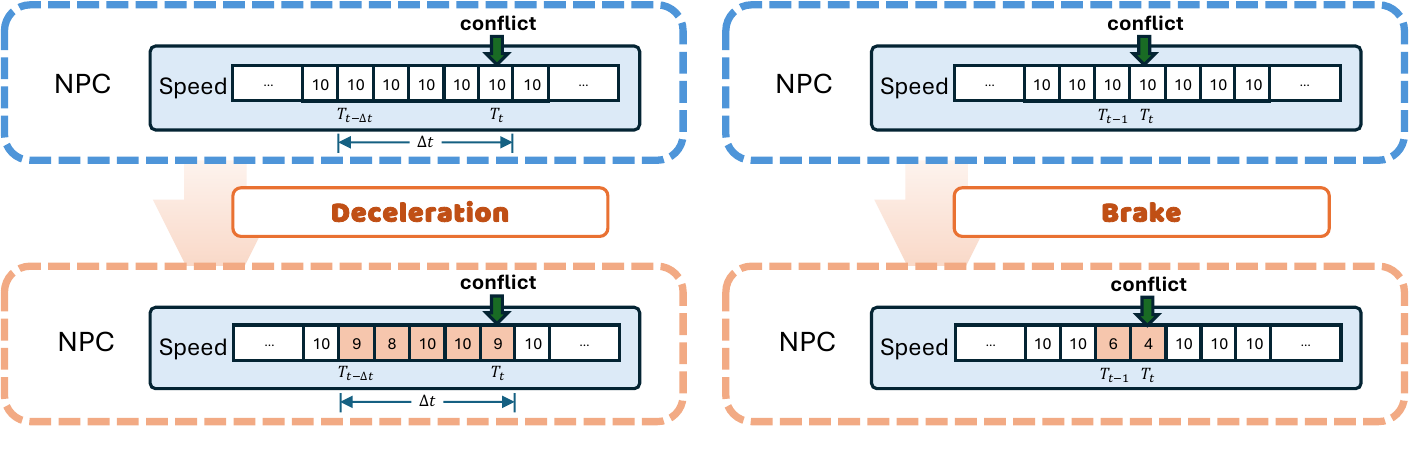}
    \caption{Examples of Deceleration and Brake mutations. In both examples, the NPC reaches the conflict space at time $t$.}
    \label{fig:mut_dec_bak}
\end{figure}

Operationally, collision search resembles conflict search but begins from the most conflict-rich scenario instead of a randomly initialized population. During each mutation, collision search first identifies the conflict type. Specifically, we easily distinguish the OP conflicts introduced in Section~\ref{back:conflict} from other conflict types. In OP conflict scenarios, the EV and NPC occupy the same lane; if the EV is ahead of the NPC, any resulting collision would most likely be due to the NPC’s actions rather than the EV's. Therefore, for an OP conflict, the following mutation operations apply exclusively when the NPC is ahead of the EV:

\begin{enumerate}
    \item \textbf{Deceleration:} Speeds from $t-{\Delta t}$ to $t$ are decreased by 0 to 2 m/s, but do not fall below 0 m/s. Figure~\ref{fig:mut_dec_bak} shows an example of Deceleration Mutation, where the NPC’s speed decreases by a random value within 0 to 2 m/s from time $t-{\Delta t}$ to $t$. $\Delta t$ is the conflict time of the conflict.

    \item \textbf{Brake:} Speeds from $t-1$ to $t$ are reduced by 2 to 6 m/s. Figure~\ref{fig:mut_dec_bak} shows an example of Brake Mutation, where the NPC’s speed decreases by a random value within 2 to 6 m/s from time $t-1$ to $t$.
\end{enumerate}

\begin{figure}[h]
    \centering
    \includegraphics[width=0.5\linewidth]{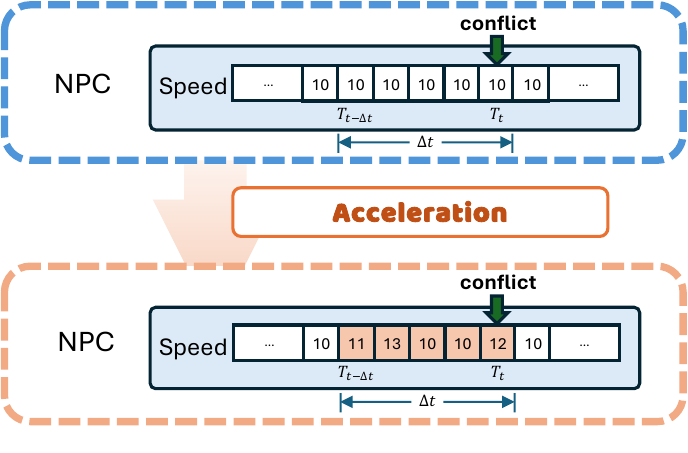}
    \caption{An example of Acceleration mutation. In the example, the NPC reaches the conflict space at time $t$.}
    \label{fig:mut_acc}
\end{figure}

For other conflict types, if the NPC reaches the conflict space first, the mutation operations are the same as those for OP conflicts. If the EV reaches the conflict space first, the mutation operation is:
\begin{enumerate}
    \item \textbf{Acceleration:} Speeds from $t-{\Delta t}$ to $t$ are increased by 0 to 3 m/s, but do not exceed the speed limit. Figure~\ref{fig:mut_acc} shows an example of \textbf{Acceleration Mutation}, where the NPC’s speed increases by a random value within 0 to 3 m/s from time $t-{\Delta t}$ to $t$.
\end{enumerate}

Similar to prior methods~\cite{li2020av,tian2022mosat,kim2022drivefuzz}, collision search uses a fitness function to gauge how close a scenario is to producing a collision. Since shorter conflict times indicate a higher likelihood of a collision, the fitness function is designed based on two key factors: the average conflict time of the conflicts and the conflict time of the most dangerous conflict. We define the fitness function as:
\begin{equation}
F_{collision} = \frac{1}{N} \sum_{i = 1}^{N} (t_c - \Delta t_{i}) + (t_c - \Delta t_{min}), 0 < \Delta t_{i} \leq t_c,
\end{equation}
where $N$ is the number of conflicts in the scenario, $\Delta t_{i}$ is the conflict time of $ith$ conflict and $\Delta t_{min}$ is the smallest conflict time among all the conflicts. This fitness function guides the collision search process to challenge the EV with near-collision conflicts to expose safety violations of the EV.

\section{Experiments}
\label{exp}
In this section, we evaluate our framework's ability to test Baidu Apollo~\cite{ap}. We conduct a comparative study against three state-of-the-art baselines: AV-Fuzzer~\cite{li2020av}, MOSAT~\cite{tian2022mosat} and Legend~\cite{tang2024legend}. To ensure a fair comparison, our evaluation is split into two parts. First, we conduct experiments on a fixed straight-road scenario to compare against AV-Fuzzer and MOSAT, accommodating their design focus for straight road scenarios. Second, to assess generalizability, we compare against Legend. In this experiment, Legend is run using its standard top-down approach starting from accident reports, while our method begins its search from a diverse set of unconstrained, randomly generated initial scenarios. Our experiments are designed to answer the following research questions:

\begin{itemize}
  \item \textbf{RQ1}: How effective and efficient is our method compared to AV-Fuzzer and MOSAT in a fixed, straight-road scenario?
  \item \textbf{RQ2}: How effective is our method at finding diverse collisions compared to Legend in unconstrained scenarios?
  \item \textbf{RQ3}: How effective is conflict search at improving the discovery of safety-critical collision scenarios?
  \item \textbf{RQ4}: Are our specialized conflict-targeted mutation operations more effective than random modifications?
\end{itemize}

\subsection{Experiment Design}
\label{exp:design}

To anser RQ1, consistent with their original experiment settings, we use the San Francisco map provided by LGSVL and select its longest straight road for a fixed, controlled scenario. Each test begins with a scenario containing the EV and two NPCs, which are all placed at predefined locations to ensure potential interaction. The initial population for the search algorithms is formed by randomly initializing the NPC chromosomes introduced in Section~\ref{method:scenario}. During the simulation of the scenarios, the NPCs will attempt to follow their chromosomes as closely as possible but will reject illegal actions, such as changing to the right lane while already in the rightmost lane. A collision listener on the EV detects any collisions that involve the EV. AV-Fuzzer, MOSAT and our method is then executed to search for collision scenarios caused by the EV and, based on the results, all collision scenarios are categorized into distinct types according to vehicle trajectories and actions. We compare the effectiveness and efficiency of each method using the following metrics:
\begin{itemize}
  \item The total number of collision scenarios. 
  \item The number of distinct types of collision scenarios.
  \item The number of search steps required to detect the first collision scenario.
  \item The average number of search steps required to detect a single collision scenario.
  \item The total number of search steps required to detect all collision types.
\end{itemize}

For the parameters, We set the conflict time limit ($t_c$) to 3 seconds, ensuring these conflicts remain strongly linked to collisions. In addition, we introduce a spatial conflict time limit ($t_s$) of 15 seconds, allowing for a larger time gap to identify spatial conflicts. We also conduct parameter analysis of the two time limit and show the results in Appendix A.1. In our search algorithms, both mutation and crossover probability thresholds in conflict search are 0.4, while the mutation probability threshold in collision search is 0.8. For every 5 iterations of conflict search, the most critical scenario will be selected and we run 5 iterations of collision search based on this scenario. Each experiment runs for 1,600 searches and is repeated three times to evaluate performance consistently.

To answer RQ2, we conduct experiment on the full San Francisco map, removing the straight-road constraint from RQ1. For this experiment, Legend is executed using its standard methodology. In contrast, our method begins with an ego vehicle initialized at \textbf{random} locations, surrounded by three NPCs with randomly generated chromosomes. The performance of both approaches is then evaluated using the same metrics outlined for RQ1.

To answer RQ3, we conduct an ablation study to isolate and measure the contribution of our initial exploration stage using the experiment setting for RQ1. We create an ablated version of our method which bypasses the conflict search stage entirely and feeds randomly generated scenarios as the initial population into the collision search stage. By comparing the performance of our full method against this variant, we can quantify the direct impact of the conflict search component on improving the diversity of collisions found.

To answer RQ4, we conduct a second ablation study to isolate the effectiveness of our atom-level conflict-targeted mutations. To create a direct comparison against the random mutation strategies applicable to prior work, we designed an experiment that neutralizes the effect of our conflict search stage. Specifically, we build another variant similar to the one for RQ3, but replace our conflict-targeted mutation strategies with random mutations like Speed Mutation and Action Mutation introduced in Figure~\ref{fig:mut_spd_act}. By comparing the performance of these two variants, we can measure the direct impact of our mutation strategy without the influence of our two-stage framework.

All experiments were conducted on a system running Ubuntu 24.04.1 with an Intel i7-9700K CPU, an Nvidia RTX 2070S GPU, and 32 GB of memory. Same as the baselines~\cite{li2020av,tian2022mosat,tang2024legend}, we used LGSVL 2021.3~\cite{rong2020lgsvl} as the simulator, with Apollo 7.0 serving as the test ADS.

\subsection{Results for RQ1}
\label{exp:rq1}

\begin{figure}[h]
    \centering
    \includegraphics[width=0.6\linewidth]{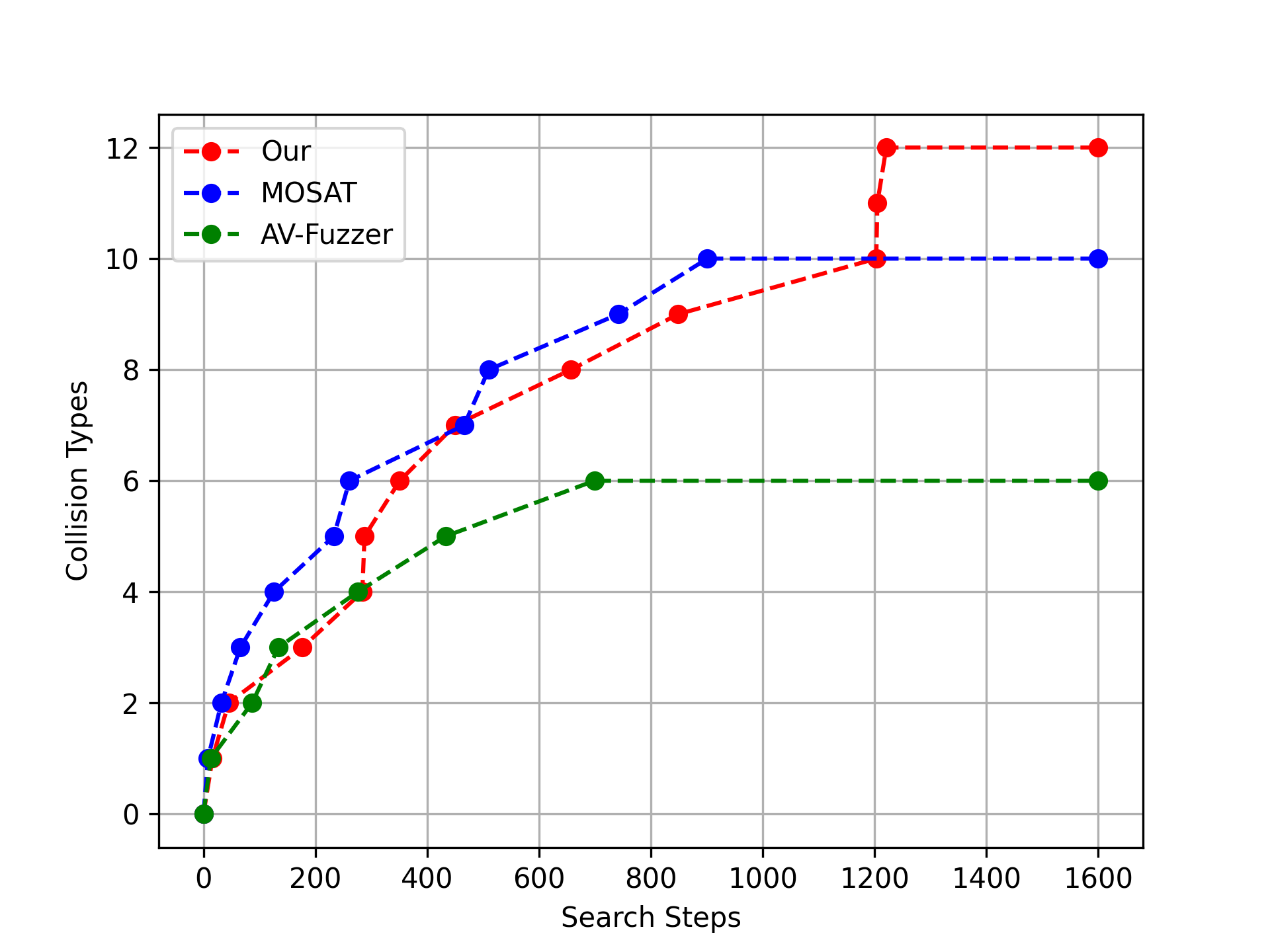}
    \caption{Growth in the number of collision types discovered by each of the three approaches, including the result in which the greatest variety of conflict types is identified.}
    \label{fig:collision_growth_s}
\end{figure}

Figure~\ref{fig:collision_growth_s} presents the growth in the number of distinct collision types discovered by each method over 1,600 search steps. MOSAT exhibits the fastest initial discovery rate, which can be attributed to its powerful, well-designed motif-based mutations. These mutations can significantly alter scenarios in a reasonable way, allowing the search to partially overcome the diversity limitations of its proximity-based fitness function. AV-Fuzzer also converges quickly but to a much smaller set of 6 types. In contrast, while our method's initial rate is comparable, it demonstrates superior \textbf{sustained exploration}, continuing to uncover new, rare collision types late in the search process after both MOSAT and AV-Fuzzer have plateaued. Ultimately, our approach discovers the most diverse set of failures, demonstrating its ability to avoid the premature convergence that limits the other methods.

To analyze the quantitative performance, we first detail the results of our method across the three repeated runs in Table~\ref{tab:result_our}. Our approach consistently uncovers a large and diverse set of safety-critical collisions, revealing between 9 and 12 distinct collision types in each run (for an average of 11). The method generates a substantial number of collisions (310 on average) and is efficient at initial discovery, requiring an average of only 12 steps to find the first failure. 

\begin{table}[h]
    \centering
    \caption{Detailed results of our method over three runs in a fixed, straight-road scenario.}
    \label{tab:result_our}
    \begin{tabular}{c|ccc|c}
        \toprule
        Metrics & Run 1 & Run 2 & Run 3 & Average \\
        \midrule
        Total Collisions & 289 & 324 & 317 & 310 \\
        Collision Types & 9 & 12 & 11 & 11 \\
        Search Steps for First Collision & 11 & 16 & 8 & 12 \\
        \bottomrule
    \end{tabular}
\end{table}

Building on these results, Table~\ref{tab:result_straight} presents the direct comparison against the baselines, showing the average performance across all runs.

\begin{table}[h]
    \centering
    \caption{Average performance comparison against AV-Fuzzer and MOSAT in a fixed, straight-road scenario.}
    \label{tab:result_straight}
    \begin{tabular}{c|ccc}
        \toprule
        Metrics & Ours & AV-Fuzzer & MOSAT \\
        \midrule
        Total Collisions & 310 & 302 & 266 \\
        Collision Types & 11 & 5 & 9 \\
        Search Steps for First Collision & 12 & 13 & 7 \\
        Average Search Steps for One Collision & 5 & 5 & 6 \\
        Search Steps for All Collision Types & 1045 & 699 & 802 \\
        \bottomrule
    \end{tabular}
\end{table}

\begin{figure}[h]
    \centering
    \includegraphics[width=0.8\linewidth]{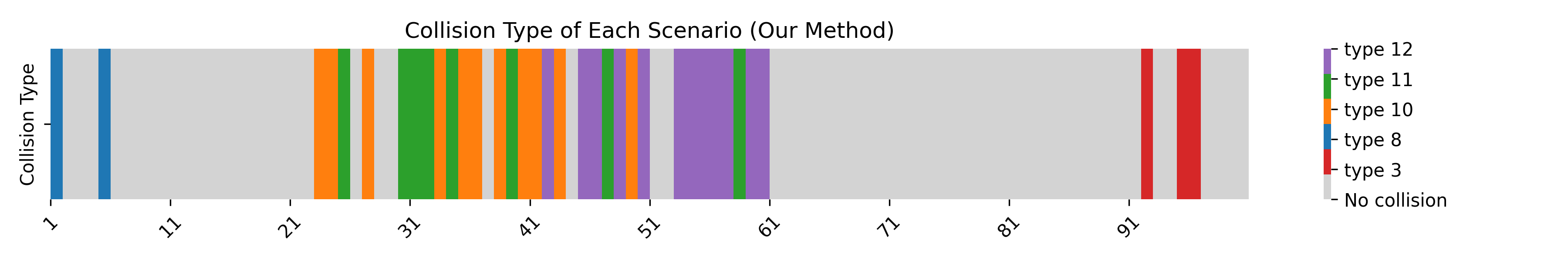}
    \includegraphics[width=0.8\linewidth]{figure/collision_heat_avfuzzer.png}
    \caption{The collision type distribution on 100 continuous search steps from our method and AV-Fuzzer. The colors are associated with different types of collision introduced in Appendix A.2.}
    \label{fig:collision type}
\end{figure}

\textbf{Ours vs. AV-Fuzzer.}
As shown in the average results in Table~\ref{tab:result_straight}, our approach is significantly more effective than AV-Fuzzer. Our method exposes an average of 11 collision types, more than double the 5 types found by AV-Fuzzer. Both methods detect their first collision at roughly the same time (12 vs. 13 steps) and find a similar total number of collisions (310 vs. 302). This indicates that for a similar search effort, our method's conflict-driven approach explores a much wider and more diverse range of collisions. To understand the reason for this dramatic difference in diversity for a similar search effort, we analyze the search behavior in detail. Figure~\ref{fig:collision type} illustrates the collision types observed over 100 consecutive search steps. While collisions in both methods occur in “groups” evolved from common parents, AV-Fuzzer's groups always belong to the same collision type. In contrast, our method is able to produce multiple collision types from same parent scenarios. This is because AV-Fuzzer's distance-based fitness function repeatedly drives the search toward the same near-collision pattern, whereas our method's use of different conflicts introduces more varied search directions.

\textbf{Ours vs. MOSAT.}
The comparison with MOSAT is more competitive and highlights the different strengths of the two approaches. As shown in Table~\ref{tab:result_straight}, MOSAT is the most efficient at finding the first collision, requiring only 7 steps on average. However, our method proves more effective on the crucial metric of overall diversity, discovering an average of 11 distinct collision types to MOSAT's 9. We attribute this to the nature of their respective mutation strategies. While MOSAT's powerful motif-based mutations enable its rapid early growth, we observe that their complexity (certain combination of atom-level mutations) also limits the variety of discoverable collision patterns, causing the search to plateau once the more common types are found. Our method's conflict-driven exploration, while requiring more steps to uncover all collision types (1045 vs. 802), is eventually more comprehensive. This demonstrates that our approach is superior for achieving a thorough safety evaluation.

A catalog of all 12 discovered collision types, each with a figure and a detailed description, is provided in Appendix A.2.

\subsection{Result for RQ2}
\label{exp:rq2}

\begin{figure}[h]
    \centering
    \includegraphics[width=0.6\linewidth]{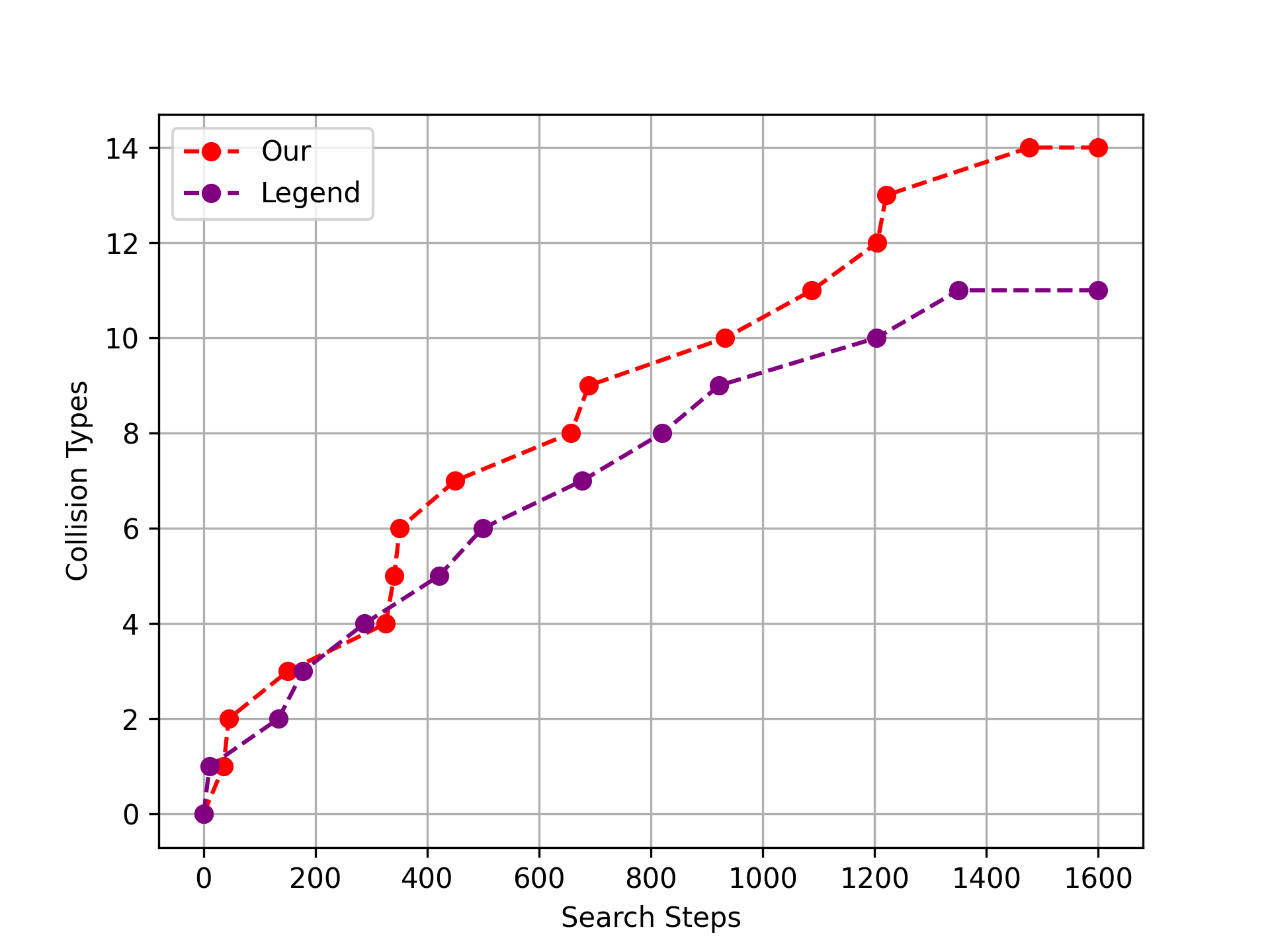}
    \caption{Growth in the number of collision types discovered by our method and Legend, including the result in which the greatest variety of conflict types is identified.}
    \label{fig:collision_growth_r}
\end{figure}

To answer RQ2, we evaluated our framework's generalizability in unconstrained scenarios against Legend, the state-of-the-art in diverse scenario generation. First, we analyze the discovery rate of collision types. Figure~\ref{fig:collision_growth_r} shows the growth in distinct types found by our method versus Legend over 1,600 search steps. Both methods demonstrate a strong ability to find diverse collisions in this more challenging environment. Legend's report-driven approach finds a steady stream of unique scenarios, reaching a total of 11 types. Our method, starting from randomly generated scenarios, keeps pace with Legend in the early and mid-stages of the search. In the later stages, our conflict-driven exploration continues to find new, rare failures after Legend's search has plateaued, ultimately discovering a greater total number of collision types.

\begin{table}[h]
    \centering
    \caption{Detailed results of our method over three runs in unconstrained scenarios.}
    \label{tab:result_our_r}
    \begin{tabular}{c|ccc|c}
        \toprule
        Metrics & Run 1 & Run 2 & Run 3 & Average \\
        \midrule
        Total Collisions & 192 & 144 & 199 & 178 \\
        Collision Types & 14 & 12 & 14 & 13 \\
        Search Steps for First Collision & 11 & 12 & 36 & 20 \\
        \bottomrule
    \end{tabular}
\end{table}

For quantitative performance, we first show the standalone performance of our method over three runs in Table~\ref{tab:result_our_r}. The results demonstrate the effectiveness of our approach in unconstrained scenarios. Even when starting from random initial scenarios, our method still can discover a high number of diverse collisions, finding between 12 and 14 distinct collision types in each run (averaging 13). This results demonstrate the effectiveness of our approach in unconstrained scenarios and validate that our search is a powerful and generalizable strategy for exploring the vast scenario space. Building on these results, Table~\ref{tab:result_straight} presents the quantitative comparison against Legend, showing the average performance across all runs.

\begin{table}[h]
    \centering
    \caption{Average performance comparison against Legend in unconstrained scenarios.}
    \label{tab:result_random}
    \begin{tabular}{c|cc}
        \toprule
        Metrics & Ours & Legend \\
        \midrule
        Total Collisions & 178 & 110 \\
        Collision Types & 13 & 11 \\
        Search Steps for First Collision & 20 & 11 \\
        Average Search Steps for One Collision & 9 & 15 \\
        Search Steps for All Collision Types & 1333 & 1356 \\
        \bottomrule
    \end{tabular}
\end{table}

\textbf{Ours vs. Legend.}
As shown in Table~\ref{tab:result_random}, our method discovers more distinct collision types than Legend (13 vs. 11) and finds substantially more total collisions (178 vs. 110). In terms of efficiency, Legend finds the initial collision faster (11 search steps compared to our 20), which may attribute to its use of formal logical scenarios translated from real-world accident reports. However, our framework achieves better long-term efficiency, requiring only 9 steps per collision on average, compared to Legend's 15 steps. This demonstrates that our exploration strategy provides a more comprehensive and efficient search over the long term. It is also important to note that while our conflict-driven framework demonstrates a clear advantage in diversity against Legend's standard approach, we position our method as a powerful alternative rather than claiming superiority. This is because Legend's effectiveness is closely linked to the diversity and quality of the real-world accident reports. Thus, Legend's performance could improve given a larger or more diverse corpus of real-world accident reports.

While most collisions are included in the collision types from RQ1, we find 4 more collision types in the experiments for RQ2. We also provide a catalog of these 4 collision types in Appendix A.3.

\subsection{Result for RQ3}
\label{exp:rq3}

\begin{table}[h]
    \centering
    \caption{Results of the experiments for RQ3 and RQ4.}\label{tab:result_conflict}
    \begin{tabular}{c|ccc}
        \toprule
         & $Ours$ & $Ours_c$ & $Ours_r$ \\
        \midrule
        Total Collisions & 310 & 197 & 123 \\
        Collision Types & 11 & 7 & 4\\
        \bottomrule
    \end{tabular}
\end{table}

In Table~\ref{tab:result_conflict}, we present our experimental results on the ablation study of conflict search and our specialized mutation operations. We refer to our full method as \textit{$Ours$}, a variant that uses only the collision search stage as \textit{$Ours_c$} (for RQ3), and a variant that uses random mutations instead of targeted ones as \textit{$Ours_r$} (for RQ4). We then compare the total number of collisions and the collision types discovered by the three methods.

\begin{table}[h]
    \centering
    \caption{Evolution of the average number of conflicts per scenario across search generations for \textit{$Ours$} and \textit{$Ours_c$}.}\label{tab:result_num}
    \begin{tabular}{c|cc}
        \toprule
        Search Generation & $Ours$ & $Ours_c$ \\
        \midrule
        Generation 0 & 3 & 3\\
        Generation 200 & 6 & 4\\
        Generation 400 & 10 & 4\\
        Generation 800 & 10 & 4\\
        Generation 1600 & 12 & 5\\
        \bottomrule
    \end{tabular}
\end{table}

The comparison between \textit{$Ours$} and \textit{$Ours_c$} in Table~\ref{tab:result_conflict} demonstrates the significant impact of the conflict search stage. Our full method finds substantially more collisions (310 vs. 197) and, more importantly, a greater diversity of collision types (11 vs. 7). This result indicates that dedicating an initial stage to exploring conflicts significantly enhances the discovery of diverse, safety-critical scenarios. Table~\ref{tab:result_num} provides further insight by showing the evolution of the average number of conflicts per scenario over generations. \textit{$Ours$} effectively increases the conflict count during the search, whereas the count for \textit{$Ours_c$} remains relatively stable. This process enriches the pool of scenarios from which collisions are potentially derived. 

\begin{figure}[h]
    \centering
    \includegraphics[width=0.6\linewidth]{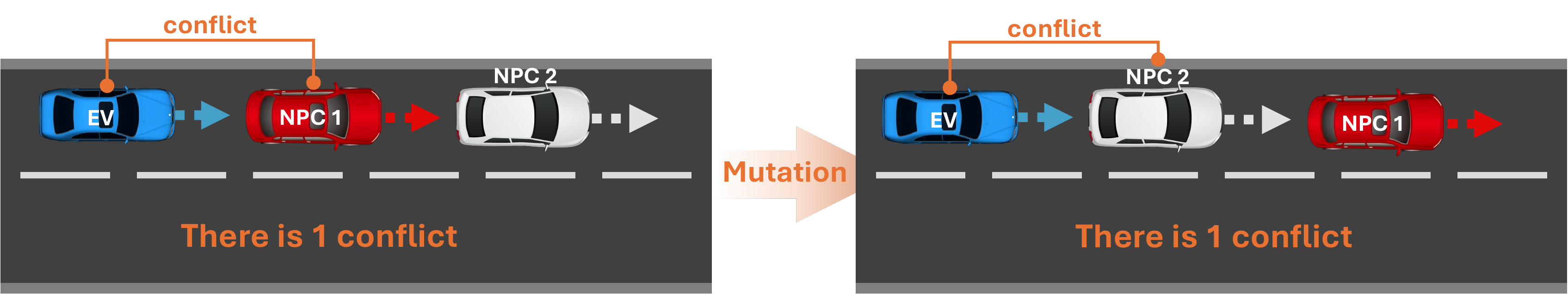}
    \caption{An example of changing conflict nature.}
    \label{fig:instance_mutation}
\end{figure}

Although the conflict count for \textit{$Ours$} plateaus between generations 400 and 800, the conflict search is still exploring new scenarios. In fact, we observe that the effectiveness of conflict search stems not only from increasing the quantity of conflicts but also from changing their fundamental nature. For instance, consider a scenario where two NPCs and the EV travel in a straight line in the order: EV, NPC1, NPC2. As illustrated in Figure~\ref{fig:instance_mutation}, the EV might initially conflict with NPC1. If we mutate NPC2’s chromosome to convert a spatial conflict with the EV into a direct conflict, the original conflict with NPC1 may resolve. Even if the total conflict count remains unchanged, this mutation introduces a new conflict interaction absent in the original scenario, potentially revealing a collision type that would have otherwise remained undiscovered.

\subsection{Result for RQ4}

To answer RQ4, we evaluate the effectiveness of our specialized conflict-targeted mutation operations. This analysis isolates the impact of the mutation strategy by comparing \textit{$Ours_c$} (which uses targeted mutations within the collision search stage) against \textit{$Ours_r$} (which replaces them with standard random mutations).

Referring back to Table~\ref{tab:result_conflict}, the results also show a significant performance difference. \textit{$Ours_c$} discovered nearly double the number of collision types compared to \textit{$Ours_r$} (7 vs. 4 types) and generated significantly more total collisions (197 vs. 123). This quantitative result demonstrates that our specialized mutations are substantially more effective at converting conflict scenarios into diverse collisions than random modifications. 

To illustrate precisely how these targeted mutations are used to enable the discovery of different collision types, we study some examples of the collision type to better explain our effectiveness. In Figure~\ref{fig:collision_growth_s}, our method exhibits a sharp increase in performance around the 1200 step, discovering three new collision types in rapid succession. We describe these three types below to understand this phenomenon.

\begin{figure}[H]
    \centering
    \includegraphics[width=0.7\linewidth]{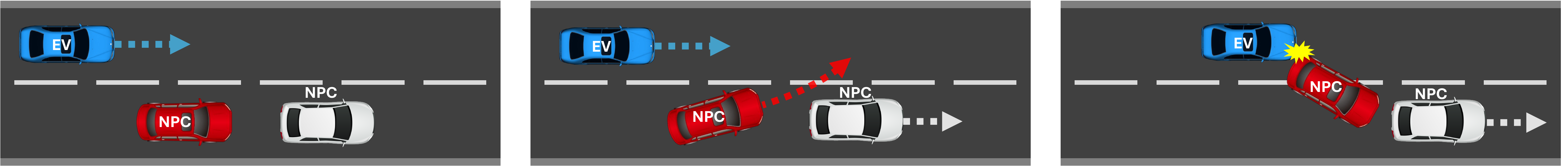}
    \caption{Example of the first type of collision scenario.}
    \label{fig:patterns_exp_1}
\end{figure}

\textbf{Example of the first type of collision scenario}: shown as Figure~\ref{fig:patterns_exp_1}, both NPC vehicles are initially in different lanes from the ego vehicle. The first NPC vehicle is stopped, while the second NPC vehicle is waiting in another lane. The second NPC vehicle then begins to merge into the ego vehicle’s lane. However, once the first NPC vehicle starts moving, the second NPC vehicle aborts its lane change and merges back. The ego vehicle fails to decelerate in time and collides with the second NPC vehicle.

\begin{figure}[H]
    \centering
    \includegraphics[width=0.7\linewidth]{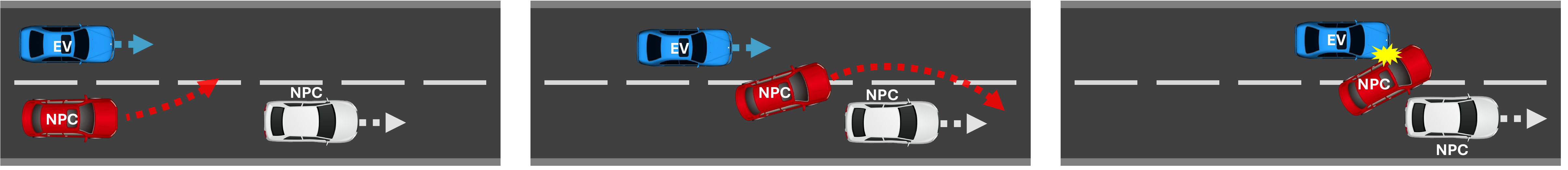}
    \caption{Example of the second type of collision scenario.}
    \label{fig:patterns_exp_2}
\end{figure}

\textbf{Example of the second type of collision scenario}: shown as Figure~\ref{fig:patterns_exp_2}, the first NPC vehicle is stopped in a different lane from the ego vehicle. The second NPC vehicle, also in a different lane from the ego vehicle, completes a lane change to overtake the first NPC vehicle but remains at a lower speed than the ego vehicle. As a result, the ego vehicle does not decelerate in time and collides with the second NPC vehicle.

\begin{figure}[H]
    \centering
    \includegraphics[width=0.7\linewidth]{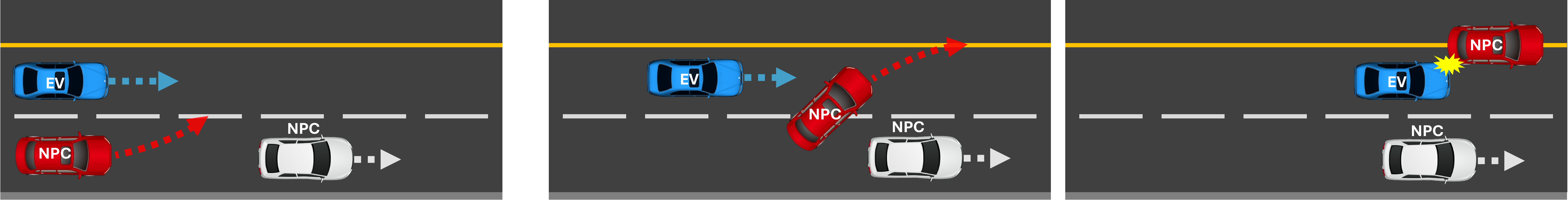}
    \caption{Example of the third type of collision scenario.}
    \label{fig:patterns_exp_3}
\end{figure}

\textbf{Example of the third type of collision scenario}: shown as Figure~\ref{fig:patterns_exp_3}, the NPC vehicle is traveling with part of its body straddling the yellow center line of the oncoming lane. As the ego vehicle passes by, it makes side contact with the NPC vehicle, resulting in a sideswipe collision.

\begin{figure}[H]
    \centering
    \includegraphics[width=0.7\linewidth]{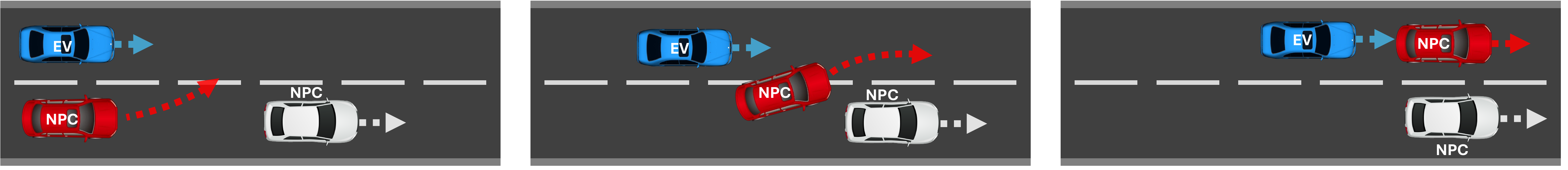}
    \caption{Common parent scenario of type 10-12.}
    \label{fig:common_patterns_10_12}
\end{figure}

These three collision types are generated from a common parent scenario, represented in Figure~\ref{fig:common_patterns_10_12}. In this scenario, the red NPC first changes lanes and then proceeds straight, creating multiple potential conflicts with the EV at different locations and times. By applying our conflict-targeted mutations to these various conflicts, the NPC is mutated to perform different actions at specific conflict timestamps. This mechanism results in different collision types being generated from the same parent scenario, which further indicates the effectiveness of our mutation strategy in leveraging conflict diversity to explore distinct collision types.
\section{Related Work}

Previous methods for ADS mainly focus on testing individual components within an ADS, including sensing\cite{broggi2013extensive,geiger2012we,jokela2019testing}, perception~\cite{tian2018deeptest,zhang2018deeproad} and planning modules~\cite{calo2020generating,wan2022too}. Some studies~\cite{boloor2020attacking,cao2021invisible,corso2019adaptive,jing2021too,li2020adaptive,sun2020towards,xu2020adversarial} employ adversarial attacks to deceive the machine learning model in sensing and perception modules, for example, Sun \etal~\cite{sun2020towards} proposes a black-box spoofing attack to generate spoofed LiDAR data to explore the general vulnerability of LiDAR-based perception architectures. Li \etal~\cite{li2020adaptive} adopts adaptive square attack to generate perturbations for traffic sign images to attack traffic sign recognition model. On the other hand, testing for planning modules focuses on assessing the safety of planned trajectories~\cite{laurent2019mutation,ohta2016pure,wan2022too}. Wan \etal~\cite{wan2022too} detects semantic Denial-of-Service vulnerabilities by introducing physical objects into driving scenes and guiding the discovery with their proposed behavioral planning vulnerability distance metric. While these methods contribute valuable insights, they often target specific components, missing broader vulnerabilities that could emerge from complex interactions among different modules.

Scenario-based testing~\cite{neurohr2020fundamental} is widely used to test the ADSs on the system level. This kind of testing employs different techniques like formal method~\cite{dreossi2019verifai,fremont2020formal,sun2022lawbreaker,zhang2023testing} and metamorphic testing~\cite{han2020metamorphic} to find critical scenarios that can reveal the system faults of the ADSs. Daniel \etal~\cite{fremont2020formal} apply formal methods to model behaviors of systems and specify the scenario to analyze the systems. Han \etal~\cite{han2020metamorphic} uses metamorphic fuzz testing to generate unrealistic scenarios to test the system robustness and recognizes genuine failures with metamorphic relations. However, techniques based on formal method and metamorphic testing still demand substantial time for sampling optimization, often resulting in similar scenarios, which limits the scalability of ADS testing in the fast-paced AV industry (e.g., Apollo, which rolls out updates weekly). 

Different from the other scenario testing methods, search-based scenario testing methods~\cite{huai2023doppelganger,kim2022drivefuzz,li2020av,lu2021search,tian2022mosat,zhong2022neural} are proved prominent for their ability to detect critical scenarios in the huge input space effectively and efficiently. AVFuzzer~\cite{li2020av} mutates
the maneuvers of NPC vehicles with genetic algorithm to find collision scenarios of the ego vehicle. MOSAT~\cite{tian2022mosat} introduces motif patterns consisting of atomic driving actions into the mutations and utilize a multi-objective function to generate diverse testing cases. Besides, DriveFuzz~\cite{kim2022drivefuzz} extends the scope of safety violations and build novel driving test oracles based on the real-world traffic rules to detect misbehaviors of the ADSs.

This paper focus on a similar task as AVFuzzer and MOSAT, which mainly explores collision scenarios. Unlike the two works, our method adds conflicts as an alternative search target, exposing various collision scenarios based on an exploration of conflicts. As shown in Section~\ref{exp}, utilizing conflicts enables our method to explore diverse safety-critical collision scenarios more effectively and efficiently.

\section{Discussion}
\label{discussion}

In Section~\ref{back:conflict}, we introduce five types of conflicts, and our method separates the OP conflict from the others. although this design is effective for our method, identifying all five conflict types and designing specialized mutation operations for each could further diversify collision scenarios, at the expense of increased design complexity and computational cost. On the other hand, our current mutations for both conflicts and spatial conflicts are based on fixing the conflict space and reducing the conflict time. We believe that additional mutations, such as actions that maintain the conflict time while reducing the distance between vehicles, could further enhance effectiveness.

Our method is currently developed on the LGSVL simulator~\cite{rong2020lgsvl} and utilizes Baidu Apollo~\cite{ap} as the testing target. However, our approach is not dependent on any specific simulator or ADS. It can be implemented on other simulators (e.g., Carla~\cite{dosovitskiy2017carla}) and used to test other ADS systems (e.g., Autoware~\cite{kato2018autoware}), as long as the simulator supports connectivity with the ADS.

\textbf{Threats to Construct Validity:} A potential threat to validity lies in the classification of discovered collision types. The process of categorizing collision types relies on qualitative observation rather than quantitative metrics. To mitigate this subjectivity, we based our classification on clearly observable differences in vehicle trajectories and actions, which we argue is reasonable as these factors directly relate to different underlying causes of the collisions. Furthermore, we provide a video catalog of all distinct collision types in Appendix A.4, following common practice in the field. 

\textbf{Threats to External Validity:} Another threat to validity is our selection of baselines (AV-Fuzzer, MOSAT and Legend). While numerous search-based scenario testing methods exist, to the best of our knowledge, these baselines have demonstrated high effectiveness and efficiency in finding diverse collisions, making them strong and relevant benchmarks. We believe that demonstrating advantages over these baselines provides a strong and valid assessment of our framework's performance.
\section{Conclusion}

In this paper, we introduce a novel search-based scenario testing framework that leverages conflicts to identify a broader range of safety-critical collision scenarios. Our approach uses a genetic algorithm to detect conflicts and employs mutation-based fuzzing to induce potential collision scenarios. We evaluate our method on Baidu Apollo and compared its performance against baseline techniques. Our experimental results show that our framework generates a significantly greater variety of collision scenarios than the baselines. Our method is also proven with the ability to reveal multiple collision types within a single scenario. Overall, our work demonstrates that conflict-based strategies can significantly enhance the effectiveness of scenario testing for autonomous driving systems, offering a promising direction for future research.

\clearpage
\bibliographystyle{IEEEtran}

\bibliography{main}

\end{document}